# ACE-RLHF: Automated Code Evaluation and Socratic Feedback Generation Tool using Large Language Models and Reinforcement Learning with Human Feedback

Tasnia Rahman[1,†], Sathish A. P. Kumar[1,‡], Sumit Jha[2,§], and Arvind Ramanathan[3,¶]

[1]Department of Computer Science, Cleveland State University, Cleveland, OH, USA
[2]School of Computing and Information Sciences, Florida International University, Miami, FL, USA
[3]Data Science and Learning Division, Argonne National Laboratory, Lemont, IL, USA
[†]t.rahman82@vikes.csuohio.edu, [‡]s.kumar13@csuohio.edu, [§]sjha@fiu.edu,
[¶]ramanathana@anl.gov

*Abstract*—Automated Program Repair tools are developed for generating feedback and suggesting a repair method for erroneous code. State of the art (SOTA) code repair methods rely on data-driven approaches and often fail to deliver solution for complicated programming questions. To interpret the natural language of unprecedented programming problems, using Large Language Models (LLMs) for code-feedback generation is crucial. LLMs generate more comprehensible feedback than compiler-generated error messages, and Reinforcement Learning with Human Feedback (RLHF) further enhances quality by integrating human-in-the-loop which helps novice students to lean programming from scratch interactively. We are applying RLHF fine-tuning technique for an expected Socratic response such as a question with hint to solve the programming issue. We are proposing code feedback generation tool by fine-tuning LLM with RLHF, Automated Code Evaluation with RLHF (ACE-RLHF), combining two open-source LLM models with two different SOTA optimization techniques. The quality of feedback is evaluated on two benchmark datasets containing basic and competition-level programming questions where the later is proposed by us. We achieved 2-5% higher accuracy than RL-free SOTA techniques using Llama-3-7B-Proximal-policy optimization in automated evaluation and similar or slightly higher accuracy compared to reward model-free RL with AI Feedback (RLAIF). We achieved almost 40% higher accuracy with GPT-3.5 Best-of-n optimization while performing manual evaluation.

## I. INTRODUCTION

LARGE Language Models (LLMs) are considered powerful tools for diversified applications. Our goal is to develop LLMs which can be used in the classroom to help the students navigate intricate programming problems. However, LLMs suffer from hallucinations and render solutions that are factually incorrect [1]. In solving programming problems this can lead the student to a rabbit hole impeding the learning process. To make a solution to this problem, fine-tuning LLMs involving a Human-in-the-loop approach improves the quality of code feedback generation. Providing human feedback directly to the LLM is impractical so we create a separate model named the reward model which adheres to human preference for a specific problem solution and provides rewards that allow a Reinforcement Learning (RL) agent to align its behavior with human-like expectations and values while training the model. RL combined with the human feedback (RLHF) technique can improve LLM alignment as well as human preferences for specific tasks and goals. In this research, we explore whether fine-tuning with RLHF enhances the accuracy of code feedback and adopting human-in-the-loop through RLHF while building LLM-based tools compared to other SOTA techniques of code feedback generation. Our goal is to establish active learning within the classroom for students who are learning programming language while providing a feedback question with hint to solve the issue within their buggy code by themselves.

Autonomous program repair (APR) tools [2] [3] have been in use for the last few years although these tools heavily rely on predefined databases whereas LLMs associate control-flow structures with the logical intent of the user and are capable of solving or repairing basic level of programming questions almost accurately with some errors for complicated program problems [4]. However, we intend to generate concise and helpful feedback for the corresponding code rather than providing the repaired code to the student. The zero-shot code generation capability of LLMs pushes for giving out direct answers to students that might not be correct. Besides, LLMs are prone to jump toward solutions that have the potential to impede student learning. Moreover, these models are somewhat prone to hallucination where they confidently provide answers which are factually wrong, logically inconsistent, and fabricated. This can be mitigated by fine-tuning the base LLM models properly. Hence, this paper takes a step towards providing question feedback with a Socratic method that can be effectively used in classes to improve students' programming learning outcomes by fine-tuning base LLMs with RLHF.

We use base LLM and fine-tune with RLHF for automated code evaluation and provide hints for repair by catapulting a question for the corresponding error in each turn of the user input code. For evaluating the efficacy of our ACE-RLHF tools, we implemented four chatbots with four corresponding fine-tuned models for performing manual evaluation where text questions and code snippets can be added for the query. We propose a dyadic conversational benchmark dataset for evaluating feedback on competition-level programming questions. Competition-level programming problems are collected from CodeForces [1] and ICPC [2]. This is the only available benchmark with a Socratic conversational scheme for intricate competition-level questions to our knowledge. The Socratic conversational turn-by-turn scheme requires extensive fine-tuning to facilitate naturalistic, context-aware feedback that addresses both logical consistency and relevance of corresponding code issues, especially when multiple turns of user interaction occur. The contributions of our research involve combining several SOTA techniques to fine-tune ACE-RLHF for better accuracy. The contributions are as follows,

- Improving Code Feedback quality using proposed ACE-RLHF while training a reward model with a Socratic preference dataset.
- Implement student-instructor Socratic conversational benchmark dataset for competition-level programming problems.
- Comparing the alignment performance of proposed ACE-RLHF, with the existing fine-tuned LLMs, and analyze the correlation between expected calibration error of reward model with tools' alignment performance.

## II. RELATED WORK

Recent advances in LLM have made the interpretation of compiler-generated error messages less complicated and easy to understand for students [5] [6] [7]. Leinonen et. al. proposed the usage of LLMs for interpreting complex compiler-generated messages for an erroneous code [8]. The source code used for this experiment was relatively small and failed to capture the capability of LLM to generate error messages.

The use of the Socratic method in debugging code errors, has been proposed by [9]. They proposed a conversational benchmark dataset[3] for basic programming questions and compared the LLM-generated response with their standard dataset using similarity-based evaluation metrics. The approach relies on fine-tuning the model with limited program metadata and prompting. In this paper, we are training a reward model with a preference dataset and feeding the scores to the base LLM model to extract the most accurate feedback for basic and complicated programming issues.

Zeng. et. al. proposes a reward model MORE and shown that there exists a correlation between the calibration performance of reward models with LLM's alignment to the expected reference response [10]. Our idea to measure the quality of feedback with alignment of the LLM-generated response and calibration error of the ACE-RLHF tool correlation, is inspired by theirs, although our domain and application are completely different.

Models that are trained on a comparatively small number of parameters and computationally less heavy are typically considered as low parameter models. However, decent performance accuracy can be achieved with these low-parameter and open-source models which requires less computational resources, operational costs and ensures accessibility to students. Hence, we build our work on base models of Llama-3, recently released from Meta AI and GPT-3.5-turbo from OpenAI, considering the balance between accuracy and computational efficiency. In this research, we have referred the gpt-3.5-turbo as GPT-3.5 for maintaining simplicity.

## III. LLM IN CS EDUCATION

Large language modeling has become a prominent area in recent times for educational activities in regard to its capability of question answering, code generation, and reasoning. These models are capable of solving intricate programming problems almost accurately. Although this attribute of LLMs has the potential to solve programming problems, it fails to provide personalized feedback depending on the student submission when the model is not fine-tuned with appropriate data and hyperparameter. Fine-tuning the LLMs with the Socratic method for question generation for buggy code can be considered as a learning material for students which can help them in solving weekly assignments and preparing for exams whereas it will help to build students' capability of reasoning and building logic to solve a problem. Moreover, sometimes it becomes hard for Teaching Assistants to provide personalized support for each student within a large class and for the primary instructor as well. Using LLMs with RLHF within this zonal proximal development can be a solution to help computer science students.

Our work promotes active learning in a classroom where students' engagement is significant. LLM can enhance the learning experience while providing helpful support yet encouraging involvement to solve a coding problem. We ensure the fine-tuned LLM model used, does not reveal the direct answer to the question. Answers that are direct and premature are eliminated at the first step of our fine-tuning process and explained in detail in the methodology section.

## IV. METHODOLOGY

The detailed methods of our research are briefly described in the following sections. Figure 1 shows the overall workflow diagram of the ACE-RLHF using the Socratic method. Our contributions on top of the SOTA RL-free techniques are depicted in green. First, the student's input code is prompted to the fine-tuned LLM model for evaluation and instruction tuned to provide feedback in a Socratic manner, and the best response is generated by optimization using RL utilizing the reward model trained on the preference dataset which consists of both valid and invalid responses. Then model response is provided to the student and it is again the student's turn to provide input on his/her query. The ACE-RLHF tools' web-based implementation is shown in Appendix B. The students were asked to use four fine-tuned models within this chat interface and log their judgment depending on the manual evaluation metrics and benchmark datasets which was provided for evaluation as well. The details methodology is provided in the following sections.

---

[1] https://codeforces.com/
[2] https://icpc.global/
[3] https://github.com/taisazero/socratic-debugging-benchmark

### A. Benchmark Dataset

Al-Hossami et. al. proposed a dataset that includes basic programming questions, possible common bugs, bug descriptions, its fix, and questions with bug fix hints [9]. The dataset is tailored in such a way that a standard response by an instructor is stored for the student's buggy code and his question regarding the issue. It has one main response for a question and several alternate responses. Responses are collected for 38 basic programming questions and common bugs related to these questions including syntactical and logical issues within 151 dialogue threads. We are using this dataset as the benchmark for basic-level programming questions for evaluating the performance of ACE-RLHF tools. We are introducing a benchmark dataset similar to theirs with the same structure as the problem statement, possible common bugs, and its fix for competition-level questions. We are collecting programming questions from CodeForces [11] and the ICPC website to create our benchmark. An example of the "Splitting Apples" problem from the basic-level benchmark dataset is shown in Illustration. 1 and the "Find the Bone" problem from our proposed Benchmark consisting of Competition-level Programming Question-Answer pair is shoiwn in Illustration. 2

---

**Illustration 1**

[breakable] Problem Name : Splitting Apples
(Input, Output, Buggy Code, Unit test cases, Bug Description, Bug Fix)

```
1.  def split_apples(apples, children):
2.      i = 0
3.      while apples > 0 and apples >
   children:
4.          apples = apples - children
5.          i += 1
6.      return i
```

**Student:**
Hi! My function fails two test cases and I do not understand why, can you help?

**Main responses:**
[1] Sure. Do you see anything special about the test cases it fails, compared to the ones where it works well?

**Alternate responses:**
[1] Sure, I can help. Let's consider the third test case where 'apples = 100' and 'children = 100'. Can you manually run through the code and explain what happens, line by line.

---

**Illustration 2**

[breakable] Problem Name : Find the Bone (796B)
(Input, Output, Buggy Code, Unit test cases, Bug Description, Bug Fix)

```
1.  def find_bone_position(n, m, k, holes,
   swaps):
2.      bone_position = 1
3.      for u, v in swaps:
4.          if bone_position == u:
5.              bone_position = v
6.          elif bone_position == v:
7.              bone_position = u
8.      return bone_position
```

**Student:**
My code isn't working. It doesn't handle the bone falling into a hole early. Can you help me find what's wrong?

**Main responses:**
[1] Sure! It looks like your code is continuing to process swaps even when the bone falls into a hole. What should happen when the bone reaches a hole?

**Alternate responses:**
[1] Sure! Can you explain your code line by line?
[1] Sure! Can you check if the bone has fallen into a hole and terminate the process if it has. Can you think of where you might add that check?

---

### B. Reinforcement Learning with Human Feedback

To align with the LLM's behavior according to human values and preferences, RL has been used to fine-tune the model [12]. RLHF enables the human-in-the-loop technique to provide preferred feedback. In RLHF, LLM responses are gathered and annotated as valid or invalid. Then a reward model is built and trained with the annotated dataset. A preference dataset[4] proposed by [13] is used which consists of 2500 tuples of valid and invalid Socratic responses for the same 38 basic programming questions and their corresponding possible bugs. Here, invalid Socratic questions were generated using GPT-4 using four criteria such as irrelevant, repeated, direct, and premature. Valid questions were extracted from the benchmark dataset as the ground truth from [9] and our proposed benchmark. Invalid questions were re-evaluated by us manually to check their authenticity, as a part of negative sampling [14] where we found all generated responses to be invalid as per our four criteria. Finally, we extended the preference dataset proposed by [13] and added 90 pairs of valid-invalid feedback pair for each question and each turn for up to three turns as the benchmark dataset we generated for competition-level questions consists of three to four turns following the same approach of generating invalid response with same four criteria using GPT-4 to maintain consistency. Although the preference dataset contains less data from competition-level problems, the reward model is already trained with valid-invalid responses which enables the reward

---

[4]https://github.com/umass-ml4ed/socratic-quest-gen/tree/main/preference_data

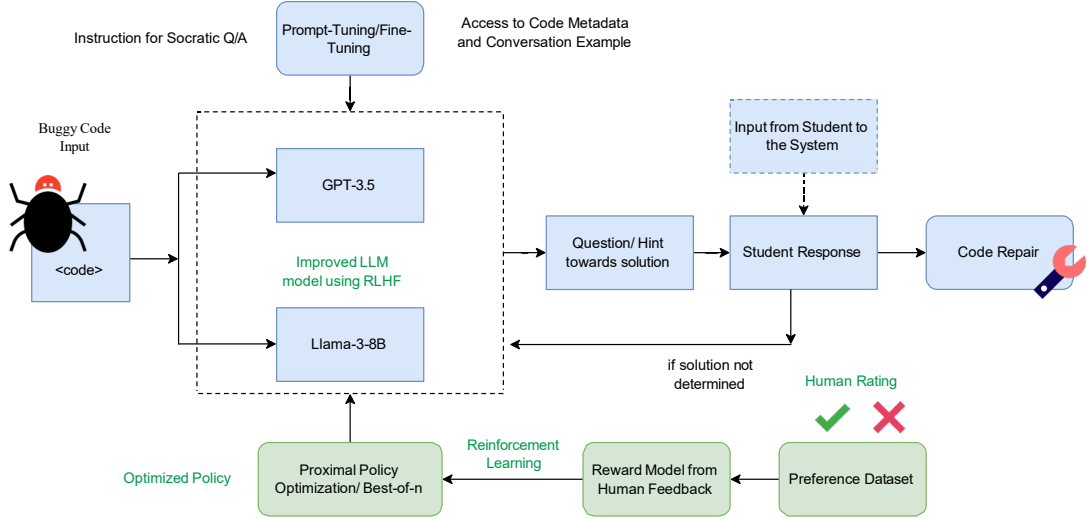

Fig. 1: RLHF framework used in this work - our modifications on top of the State-of-the-art setup for Code Evaluation and Feedback are highlighted in green

model to learn preferred Socratic output. We are implementing the widely known **Proximal Policy Optimization** algorithm and **Best-of-n** technique to optimize the ACE-RLHF using the reward model.

### C. Reward Model

The calibration performance of the reward model typically refers to the expected calibration error (ECE) of the model, on which the expected alignment performance of LLM can depend. We are training two reward models for policy optimization for LLM utterance for code feedback, using best-of-n and Proximal Preference Optimization (PPO) with the preference dataset mentioned in the previous section. We extract the reward model's accuracy and corresponding expected calibration error using equations 1-6. In equation 1, D is the preference dataset, $r_\theta$ is the preference score with $r_\theta(x, y_w) > r_\theta(x, y_l)$ that implies LLM utterance $y_w$ is preferred for the particular context. Hence, the reward model learning objective for the preference dataset $(x, y_w, y_l)$ D is defined in equation 1. Then LLM alignment is optimized for generating policy $pie(y|x)$ by maximizing the expected reward value for the LLM response using equation 2. Equation 3, shows the optimization technique typically for the PPO algorithm using reject sampling to avoid the RL schedule during the alignment process, and reject sampling loss is calculated where $y^{best} = \arg\max_{1 \leq s \leq S}\{r(x, y^s)\}$ represents the sampling response which has the highest reward score.

$$L_{rank}(\theta; D) = -E_D [\log (\sigma(\Delta r_\theta(y_w, y_l)))] \quad (1)$$

$$E_{x \sim D, y \sim \pi(y|x)} [r_\theta(x, y)] - \beta D_{KL} [\pi(y|x) \| \pi_{ref}(y|x)] \quad (2)$$

$$L_{RJS}(\pi) = -E_{x \sim D, y \sim \pi(y|x)} \log \pi(y^{best}|x) \quad (3)$$

Calibration error is a metric to analyze the confidence of a model's output. The confidence interval is considered within 0 to 1 finite samples of M bins of length 1/M and model predictions are placed into these bins depending on their prediction confidence. Here, $B_m$ denotes the set of indices of samples that fall into the internal ($\frac{m-1}{M}$, $\frac{m}{M}$). $y_i$ is the ground truth for i-th sample and $\hat{y}_i$ is the prediction such as LLM utterance for generating Socratic question for buggy code. If $\hat{y}_i$ is equal to $y_i$, I yields the value of 1, otherwise 0. $\hat{p}_i$ refers to the prediction confidence for the i-th sample and in equation 5, $\hat{p}_i$ is the $\sigma(\Delta r_\theta(y_w, y_l))$ from equation 1. Finally, the accuracy and average confidence of each bin is calculated using the following equations,

$$\text{acc}(B_m) = \frac{1}{|B_m|} \sum_{i \in B_m} I(\hat{y}_i = y_i) \quad (4)$$

$$\text{conf}(B_m) = \frac{1}{|B_m|} \sum_{i \in B_m} \hat{p}_i \quad (5)$$

We are analyzing the correlation between the reward model's Expected Calibration Error (ECE) and the LLM alignment. The following equation is being used to calculate the ECE for N samples.

$$\text{ECE} = \sum_{m=1}^{M} \frac{|B_m|}{n} |\text{acc}(B_m) - \text{conf}(B_m)| \quad (6)$$

We are using one RL-based optimization techniques, **Best-of-n** that generates n candidate responses for a given prompt, and the candidate with the highest evaluation score, such as the highest predicted reward, is selected as the best response. We are also using another widely known **Proximal Policy Optimisation** for LLM fine-tuning.

| Base Model | RL | ECE | BLEU-4 | | | Rouge-L | | | BERT F1 | | | CodeBLEU | | |
|---|---|---|---|---|---|---|---|---|---|---|---|---|---|---|
| | | | P | R | F1 | P | R | F1 | P | R | F1 | P | R | F1 |
| GPT-3.5 + CoT (SOTA) | - | - | **2.3** | 0.8 | **1.1** | 20.3 | 9.7 | 12.0 | 61.7 | 35.8 | 41.6 | - | - | - |
| DPO Greedy (SOTA) | - | - | - | - | - | 30.6 | 13.3 | 17.1 | **65.9** | 32.7 | 40.3 | - | - | - |
| DPO Sample-5 (SOTA) | - | - | - | - | - | 15.1 | **27.9** | 18.3 | 34.8 | **64.3** | 42.0 | - | - | - |
| GPT-3.5+CoT | PPO | 24.5 | 1.8 | 0.8 | 1.0 | 27.03 | 12.24 | 15.66 | 60.79 | 29.38 | 36.48 | **2.0** | **0.8** | **1.1** |
| | Best-of-n | 32.5 | 1.5 | 0.7 | 0.9 | 23.4 | 11.4 | 12.8 | 51.0 | 29.25 | 33.93 | 1.5 | 0.8 | 1.0 |
| Llama-3+CoT | PPO | 12.1 | 1.5 | 0.6 | 0.8 | **30.8** | 20.6 | **18.5** | 64.13 | 36.7 | **44.6** | 1.5 | 0.8 | 1.0 |
| | Best-of-n | 28.0 | 1.09 | 0.7 | 0.8 | 23.38 | 11.11 | 14.03 | 59.4 | 27.56 | 30.8 | 0.96 | 0.6 | 0.9 |

TABLE I: Automated Evaluation of ACE-RLHF using State-of-the-art Benchmark dataset with Basic Programming Questions.

| Base Model | RL Optimization | ECE | BLEU-4 | | | Rouge-L | | | BERT F1 | | | CodeBLEU | | |
|---|---|---|---|---|---|---|---|---|---|---|---|---|---|---|
| | | | P | R | F1 | P | R | F1 | P | R | F1 | P | R | F1 |
| GPT-3.5 +CoT | PPO | 24.5 | 3.6 | 2.3 | 2.0 | 21.3 | 15.6 | 15.8 | 60.1 | 55.2 | 55.7 | 4.9 | 3.1 | 3.5 |
| | Best-of-n | 32.5 | **6.6** | **5.2** | **4.8** | **28.14** | **23.5** | **25.5** | **66.64** | **63.4** | **64.1** | **5.5** | **4.8** | **5.0** |
| Llama-3-8B +CoT | PPO | 12.1 | 2.5 | 1.8 | 1.1 | 12.1 | 7.4 | 8.1 | 50.2 | 42.3 | 42.6 | 3.8 | 2.9 | 2.8 |
| | Best-of-n | 28.0 | 2.8 | 1.8 | 1.14 | 12.6 | 8.4 | 9.0 | 51.3 | 45.3 | 48.7 | 3.2 | 2.5 | 2.1 |

TABLE II: Automated Evaluation of ACE-RLHF using Proposed Benchmark dataset with Competition-level Programming Questions

*D. Fine-Tuning*

A reinforcement learning-based approach with human feedback is being integrated into base LLM models to improve the LLM utterance. The reward model is being trained with the preference dataset which uses reject sampling to eliminate invalid responses from the candidate outputs. Figure.2. shows the detailed workflow of the research. Base Model GPT-3.5-turbo and Llama-3-8B are provided access to the metadata from the original dataset from [9]. Then, these standard fine-tuned models are optimized using the maximum scores which are gathered by the reward model inside the RLHF framework within the diagram.

The models were instructed to provide responses in a Socratic manner and the models had access to a few examples of 38 basic and 30 competition-level programming problems, common bugs, fixes, and examples of the dyadic conversation.

After training the reward model, we apply RL using Proximal policy optimization and Best-of-n which enables the LLM to provide the best response, for the same prompts (user input to fix a buggy code), considering the highest reward score obtained from the rejection sampling optimization function. For PPO, we are using the learning rate of $5e^{-6}$, batch size of 64, and collecting results after 10 epochs and log-probability-based loss is being calculated using the reward model. For Best-of-n, we are setting the number of candidates as 5, the temperature as 0.0, the maximum number of tokens as 1024, and the probability of cut-off as 0.01.

*E. Automated Evaluation Metrics*

We are using four similarity-based evaluation metrics to calculate the accuracy of our proposed fine-tuned model. Distance between model-generated text/code and reference text/code is being calculated using Bilingual Evaluation Understudy-4-gram (BLEU-4), code-based BLEU-4 (CodeBLEU-4), Bidirectional Encoder Representations from Transformers F1 (BERT F1), and Recall-Oriented Understudy for Gisting Evaluation-Longest Common Subsequence (Rouge-L). The overlap of generated LLM utterance and reference ground truth utterance text, up to 4-grams, is calculated using BLUE-4 and CodeBLEU-4. The longest common sub-sequence between generated and reference text from the perspective of their summarization quality is evaluated by the Rouge-L. This matrix is able to capture the overall structure and content better compared to BLEU-4. Finally, BERT F1 uses the DeBERT [15] language model to generate embeddings for each token and compute similarity scores between embeddings based on their semantic similarity rather than the exact n-gram matching. This matrix has the capability to handle the paraphrased feedback utterances of two responses. A complete bipartite graph is created using the LLM utterance using the proposed fine-tuning method and benchmark utterance, where its multiplication is used to calculate the weight of each edge of the graph that is computed with the evaluation metrics for similarity check using the metrics. Edmond's Blossom algorithm [16] is being used to find the maximum matching in the bipartite graph. Finally, Precision, Recall, and F1 Score are calculated using the True Positives computed by cumulative weights of all edges discovered in the optimal matching, and False Positive is calculated by subtracting true positives from total LLM generated utterances. Also, a False negative is calculated by subtracting true positives from the total utterances of the benchmark response.

*F. Manual Evaluation Metrics*

To ensure the models' acceptability in code debugging and feedback generation, manual evaluation with students is crucial. For our research, we selected 10 undergraduate and graduate students who are Computer Science majors and work with coding problems on a regular basis. [9] used a subset of 11 problems from their dataset for manual evaluation. We are selecting the same subset for our manual evaluation of basic programming questions to compare our results with SOTA techniques. However, we are evaluating the ACE-RLHF tools' performance using all 30 competition-level programming questions. We build four ACE-RLHF tools with Streamlit and Gradio. We host the models within an A100 GPU and use port-forwarding to launch the web-based application consisting of the chatbot interface. The front-end of base GPT-3.5-turbo was designed with streamlit[5] and for Llama-3, we used Gradio[6]. The students did not have access to the knowledge of which model they were using. We numbered the models from 1 to 4 and asked them to record their responses within the manual we provided for each model. If they think the response semantically matches with any of the benchmark responses (students were provided a copy of the benchmark response for their designated code problems), they will record the first turn as "True Positive". If they think the response is invalid, will record it as "False Positive". Finally, if they think the output is missing in the benchmark responses but seems reasonable, will record it as "False Negative." This process is also carried out in [9] for manual evaluation. The same process is carried out until the student is satisfied with the models' response and able to repair his code. The students were asked to use the models for debugging 4 problems each and record their opinions about the models' output after each turn. One student evaluated 5 questions and all 9 students evaluated 4 different questions. One basic programming question from the basic benchmark and three competition-level questions from the proposed benchmark, in total 4 questions, were designated to each student (one student got 5 questions). Students did not receive any other external help for code debugging while using ACE-RLHF tools. After the extensive evaluation process, students were asked to rate the models on a scale of 1 to 10 depending on the models' relevancy, fluency, informativeness to the programming problem, capability of completing the task, and overall performance. Illustration. 3 shows the first turn conversation between the student and the ACE-RLHF tool which is fine-tuned with GPT-3.5-Best-of-n optimization in the real world (full conversation is provided in Appendix A).

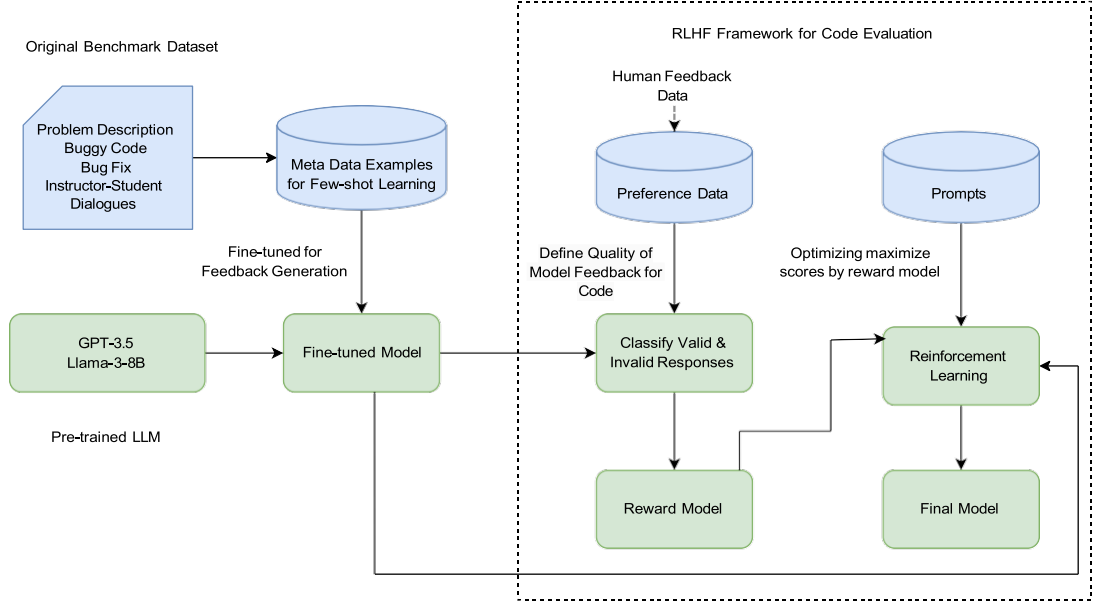

Fig. 2: Overall workflow diagram of fine-tuning LLM using Reinforcement Learning for Socratic Question Generation for Code Evaluation and Feedback

| Model | RL Optimization | Precision | Recall | F1 Score |
|---|---|---|---|---|
| **GPT-3.5 +CoT (SOTA)** | - | 18.6 | 5.5 | 8.5 |
| **GPT-4 +CoT (SOTA)** | - | 38.2 | 57.5 | 45.9 |
| **GPT-3.5 +CoT** | PPO | 69.5 | 75.5 | 72.2 |
|  | Best-of-n | **86.4** | **77.1** | **81.6** |
| **Llama-3 +CoT** | PPO | 69.7 | 36.5 | 47.8 |
|  | Best-of-n | 58.2 | 32.9 | 41.1 |

TABLE III: Manual Evaluation of ACE-RLHF using State-of-the-art Benchmark with Basic-level Programming questions

| Model | RL Optimization | Precision | Recall | F1 Score |
|---|---|---|---|---|
| **GPT-3.5 +CoT** | PPO | 71.6 | 78.3 | 74.6 |
|  | Best-of-n | **83.4** | **67.2** | **74.2** |
| **Llama-3 +CoT** | PPO | 75.6 | 70.5 | 72.1 |
|  | Best-of-n | 67.7 | 66.2 | 66.5 |

TABLE IV: Manual Evaluation of ACE-RLHF using Proposed Benchmark dataset with Competition-level Programming Questions

[5] https://streamlit.io/
[6] https://www.gradio.app/

| Model | RL Optimization | Precision | Recall | F1 Score |
|---|---|---|---|---|
| GPT-3.5 +CoT | PPO | 71.2 | 77.2 | 73.5 |
|  | Best-of-n | **84.8** | 69.4 | **75.2** |
| Llama-3 +CoT | PPO | 73.5 | 59.1 | 65.8 |
|  | Best-of-n | 66.5 | 57.4 | 61.2 |

TABLE V: Manual Evaluation of ACE-RLHF using both Basic and Competition-level Programming questions

---

**Illustration 3**

[1] Student : My code isn't working. It doesn't handle the bone falling into a hole early. Can you help me find what's wrong?

```
1. def find_bone_position(n, m, k, holes,
      swaps):
2.     bone_position = 1
3.     for u, v in swaps:
4.         if bone_position == u:
5.             bone_position = v
6.         elif bone_position == v:
7.             bone_position = u
8.     return bone_position
```

[1] Assistant: What happens if the bone falls into a hole during one of the swapping operations?

---

## V. EXPERIMENTAL RESULTS

We set the LLMs' temperature of 0.0, a greedy decoding setting for a deterministic response with Chain-of-Thought (CoT), and integrate the reward model with PPO and Best-of-n. The maximum token was set to 1024 and the probability cut-off was set to 0.01. The number of responses generated by the model is 1 for both models and the best response is generated based on the reward score and RL-based policy optimization.

Table. 1. shows the experimental results of our research and all the values are in percentage. We added the models' performance and their corresponding ECE with the RL optimization and reward model combined. The first row of the table showing the result of GPT-3.5 with CoT is the SOTA results that were collected from [9]. The second and third row shows the SOTA results from model-free RL technique with AI feedback depicted by [13] where a direct preference optimization is used without any human involvement for feedback generation. When we analyze the Precision (P), Recall (R), and F1 scores, we see Llama-3 with PPO optimization performs better than most of all SOTA techniques with precision of 30.8% and the error rate is also lowest. GPT-3.5-PPO shows better results as well where the reward model's error rate is lower than other techniques as well. This indicates somewhat correlation between the reward model's accuracy and LLM's performance. SOTA RLAIF technique, DPO Sample-5 and DPO Greedy, perform with slightly better or similar accuracy for Rouge-L recall and BERT F1 precision respectively. DPO Sample-5 performs significantly better with BERT F1 recall. For BLEU-4, SOTA results keep prevailing with slightly better accuracy as well. However, Rouge-L metric is considered to be more aligned than BLEU-4 with human judgment [17].

Therefore, we can rely on the Rouge-L scores for the proposed ACE-RLHF's performance evaluated on this metric. Besides, we achieved almost similar results using CodeBLEU metric which is capable of capturing the distance of code-based text. Bold numbers indicate the best result for the corresponding metric within the tables. As model-free SOTA results show similar and slightly higher accuracy for some metrics, manual evaluation is crucial to understand the comparative efficiency of the proposed models accurately for the target audience.

Table. 2, we applied the same assessment techniques and see ACE-RLHF tool using GPT-3.5 with CoT and Best-of-n optimization technique outperforms other models in feedback alignment for the competition-level programming question benchmark and there exits a consistency among the results using all the metrics. However, the correlation between reward models' performance and LLM alignment seems to be decreased. We further analyze our results with manual evaluation techniques described in the above section in Table. 3, and observe that GPT-3.5 with CoT and Best-of-n performs almost 40% better than the current SOTA technique with an accuracy of 81.2%. Finally, we analyze the performance of the models towards the feedback of our competition-level programming questions. We see the consistency in the result that GPT-3.5 with CoT and Best-of-n optimization performs better compared to other models. Table. 4. shows the same results but only for competition-level benchmarks. Table. 5, shows the overall performance of manual evaluation combined with the results of basic and competition-level questions.

Figure. 3 shows the results of the models' acceptability. Students rated every model's acceptability for different metrics discussed in the previous sections, within the medium to high range. For example, Figure. 3a shows 6 out of 10 students, believes each models' relevancy was within the range of 4 to 7. Figure. 3b depicts students decision about models' fluency and 9 out of 10 students rated the fluency of Llama-3 with CoT and PPO optimization to be most fluent scoring within the range of 8-10. Except for GPT-3.5 with Cot and PPO, most students rated the relevancy of the question which was asked from medium to high range. Most students preferred Llama-3 with CoT and PPO for the model's fluency. When it comes to informativeness, students chose GPT models over Llama. However, for the capability of completing the task, students chose Llama-3-CoT with the PPO model. We can see in the manual evaluation (Table. 4) of competition-level questions Llama-3 with PPO performs (72%) slightly lower than GPT-3.5 with CoT and Best-of-n (74%) which justifies the students' preferability towards the model as well.

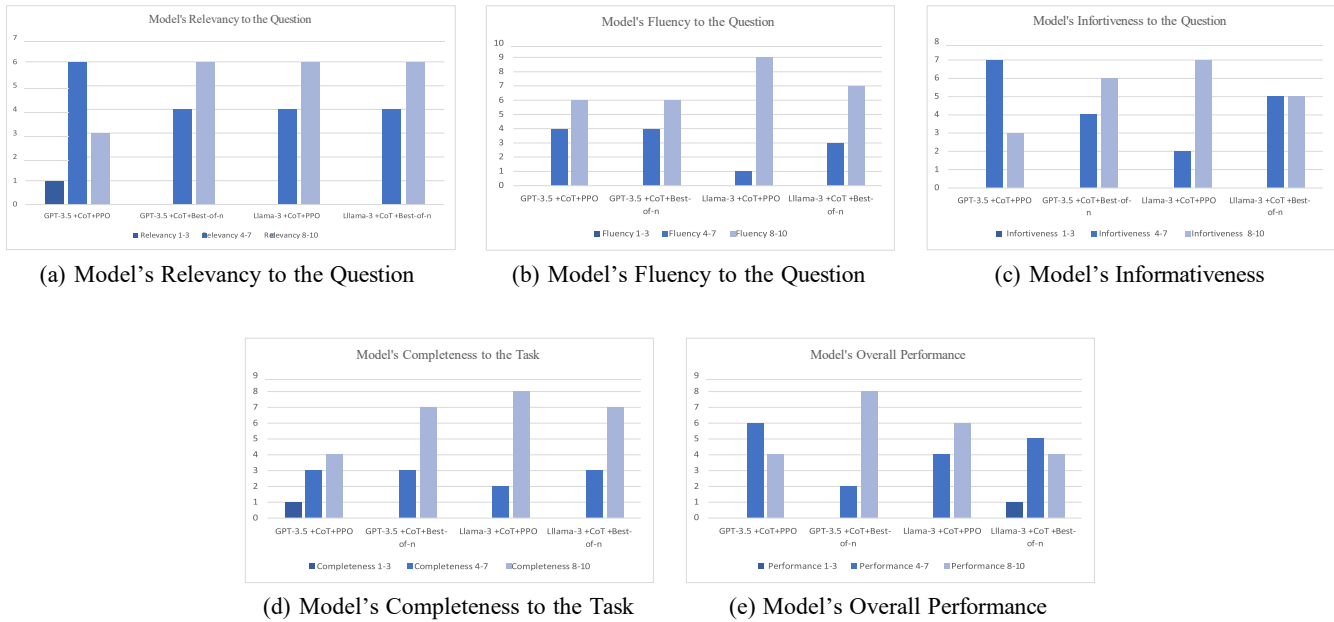

Fig. 3: Qualitative Analysis of manual evaluation of the Models

## VI. LIMITATIONS

For this research, we are using a single reward model and training with a dataset and ECE for every model is comparatively high, even though the model outperforms SOTA techniques in terms of similarity scores. LLMs often struggle with hallucination, that can be dangerous in an educational context where incorrect guidance could mislead learners. To address these issues, more complex reward models such as ensemble reward models can be explored. Another, limitation of our work is the trade-off between model complexity and accuracy. Even though we are able to achieve higher accuracy than SOTA techniques, our accuracy is almost 1-5% higher than SOTA techniques in automated evaluation of basic-level programming problems and there is a question of using such a complex architecture unless improvement is significantly higher.

## VII. CONCLUSION

We analyzed whether RL-based approaches with human-curated and preferred datasets can improve the feedback quality of the students' submitted code or not. Results show that RL-based approaches with human feedback improve code evaluation, providing high-quality feedback that promotes student engagement and active learning. From the qualitative analysis of ACE-RLHF tools, we can see students are finding the tools helpful for code debugging and tools are being rated quite well in real-world practice. Therefore, adding RLHF scheme while using LLM for code repair and feedback generation is crucial. In the future, we would also explore more baseline models and add techniques like ensemble and LLM-based reward models.

## APPENDIX A
## EXAMPLE OF SOCRATIC CONVERSATION WITH ACE-RLHF

[1] Student : My code isn't working. It doesn't handle the bone falling into a hole early. Can you help me find what's wrong?

```
1. def find_bone_position(n, m, k, holes,
     swaps):
2.     bone_position = 1
3.     for u, v in swaps:
4.         if bone_position == u:
5.             bone_position = v
6.         elif bone_position == v:
7.             bone_position = u
8.     return bone_position
```

[1] Assistant: What happens if the bone falls into a hole during one of the swapping operations?

[2] Student : I think the bone should fall into the hole and no further swaps should affect it. (using the 2nd Turn student response from benchmark)

[2] Assistant : How can you modify the code to ensure that if the bone falls into a hole during a swapping operation, it will not be affected by any future swaps?

[3] Student : I think I should add a check after each swap to see if the bone has fallen into a hole and terminate further swaps.

[3] Assistant : How can you implement this check in the code to terminate further swaps if the bone falls into a hole?

[4] Student : I checked with the following condition within my code

```
1. holes_set = set(holes)
2. if bone_position in holes_set:
3.     return bone_position
```

[4] Assistant : How can you ensure that the code checks if the bone has fallen into a hole after each swap and terminates further swaps if needed?

[5] Student : I checked with this condition and it worked.

[5] Assistant : Great job! Is there anything else you would like to discuss or any other questions you have?

[6] Student : No. Thanks!

## Appendix B
## ACE-RLHF Interface Example

Fig. 4: ACE-RLHF with GPT-3.5-Best-of-n